\documentclass[runningheads]{llncs}
\usepackage[T1]{fontenc}
\usepackage{graphicx}
\usepackage[english]{babel}
\usepackage{xcolor}
\usepackage{booktabs}
\usepackage{subcaption}
\usepackage{float}
\usepackage{microtype}
\usepackage{amsmath}

\usepackage{blindtext}

\newcommand{\todo}[1]{}
\renewcommand{\todo}[1]{{\color{red} TODO: {#1}}}

\usepackage{hyperref}
\begin{document}
\title{TextBite: A Historical Czech Document Dataset for Logical Page Segmentation}
\titlerunning{TextBite: A Dataset for Logical Page Segmentation}
% If the paper title is too long for the running head, you can set
% an abbreviated paper title here
%
% \author{Authors hidden}
% \authorrunning{Authors hidden}
\author{Martin Kostelník\inst{1}\orcidID{0009-0002-5478-9580} \and
Karel Beneš\inst{1}\orcidID{0000-0002-0805-1860} \and
Michal Hradiš\inst{1}\orcidID{0000-0002-6364-129X}}

\authorrunning{M. Kostelník et al.}
% First names are abbreviated in the running head.
% If there are more than two authors, 'et al.' is used.
%
% \institute{Institute hidden}
\institute{Faculty of Information Technology, Brno University of Technology,\\Brno, Czech Republic\\ \email{\{ikostelnik,ibenes,ihradis\}@fit.vut.cz}}
\maketitle              % typeset the header of the contribution
\begin{abstract}
    Logical page segmentation is an important step in document analysis, enabling better semantic representations, information retrieval, and text understanding. Previous approaches define logical segmentation either through text or geometric objects, relying on OCR or precise geometry. To avoid the need for OCR, we define the task purely as segmentation in the image domain. Furthermore, to ensure the evaluation remains unaffected by geometrical variations that do not impact text segmentation, we propose to use only foreground text pixels in the evaluation metric and disregard all background pixels. To support research in logical document segmentation, we introduce TextBite, a dataset of historical Czech documents spanning the 18th to 20th centuries, featuring diverse layouts from newspapers, dictionaries, and handwritten records. The dataset comprises 8,449 page images with 78,863 annotated segments of logically and thematically coherent text. We propose a set of baseline methods combining text region detection and relation prediction. The dataset, baselines and evaluation framework can be accessed at \url{https://github.com/DCGM/textbite-dataset}.
    
\end{abstract}
\section{Introduction}
    Logical page segmentation is a process of dividing a document page into smaller, meaningful, semantically coherent units. It is a crucial task in the analysis of documents, enabling the extraction of meaningful content from scanned or digitized texts. For example, such segmentation allows the creation of finer text embeddings, which can in turn improve search and retrieval results. Consequently, the work of teachers, librarians, or researchers is simplified by enabling them to find documents related to their topics of interest \cite{retrieval,liu2025passage}.
    
    Traditional approaches to logical page segmentation problem typically fall into two categories: (1) text-based segmentation, where boundaries are detected in a sequential text stream \cite{somasundaran2020two}, and (2) object-detection-based methods \cite{wang2024dlaformer,wang2024detect}, which detect paragraphs of text and subsequently connect them together to form the final segmentation. However, these methods are inherently bound to evaluation schemes, which both suffer from key limitations. Text-based approaches depend on Optical Character Recognition (OCR), tightly coupling evaluation performance with OCR accuracy. They also require a consistent reading order, which OCR may not reliably provide. On the other hand, geometric object-detection-based methods may overly penalize minor, inconsequential deviations in the shapes or sizes of annotated objects.
    \begin{figure}[ht]
        \centering
        \begin{subfigure}{0.45\textwidth}
            \centering
            \includegraphics[width=\linewidth]{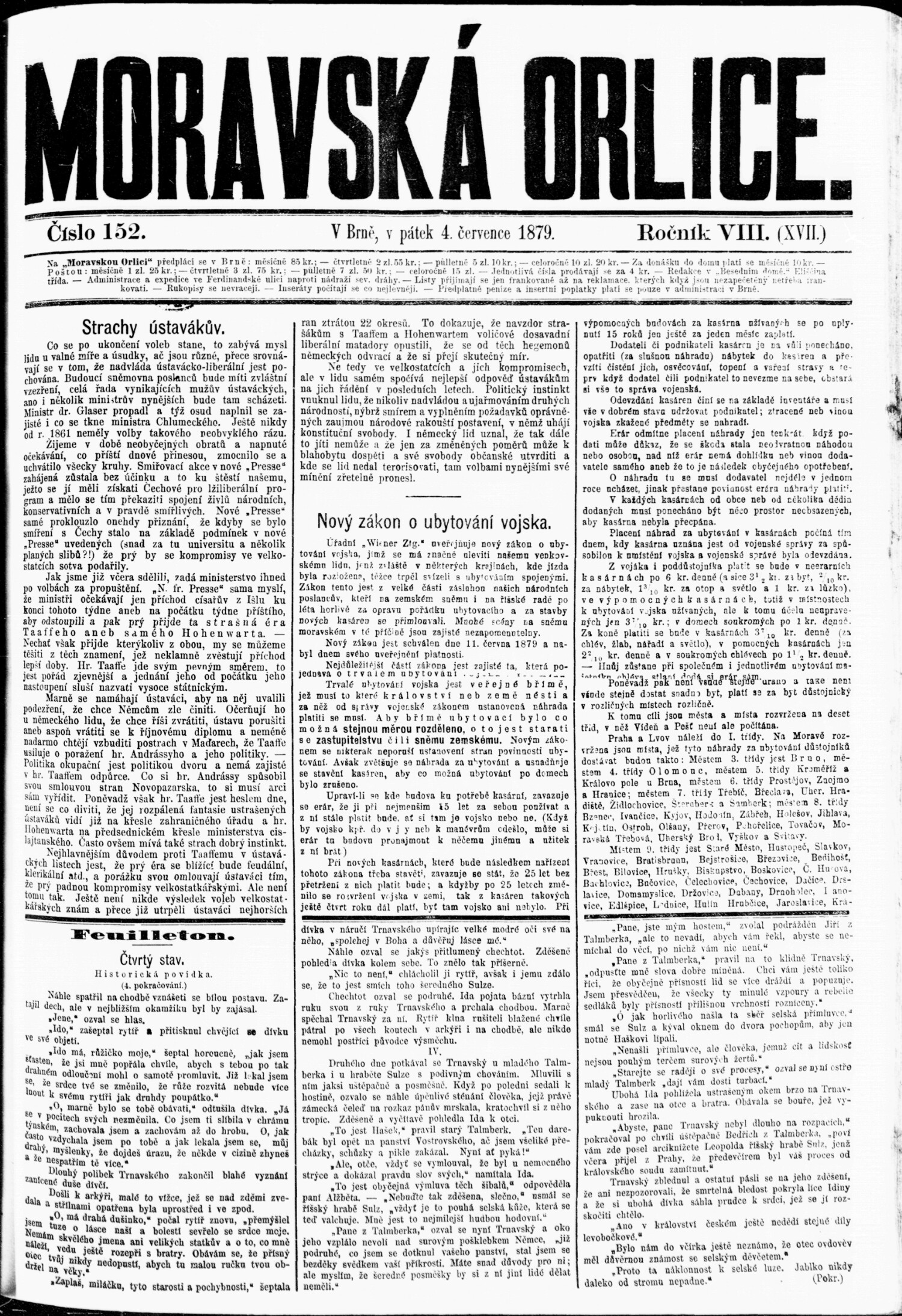}
            \caption{Input page}
            \label{fig:subfig1}
        \end{subfigure}
        \hfill
        \begin{subfigure}{0.45\textwidth}
            \centering
            \includegraphics[width=\linewidth]{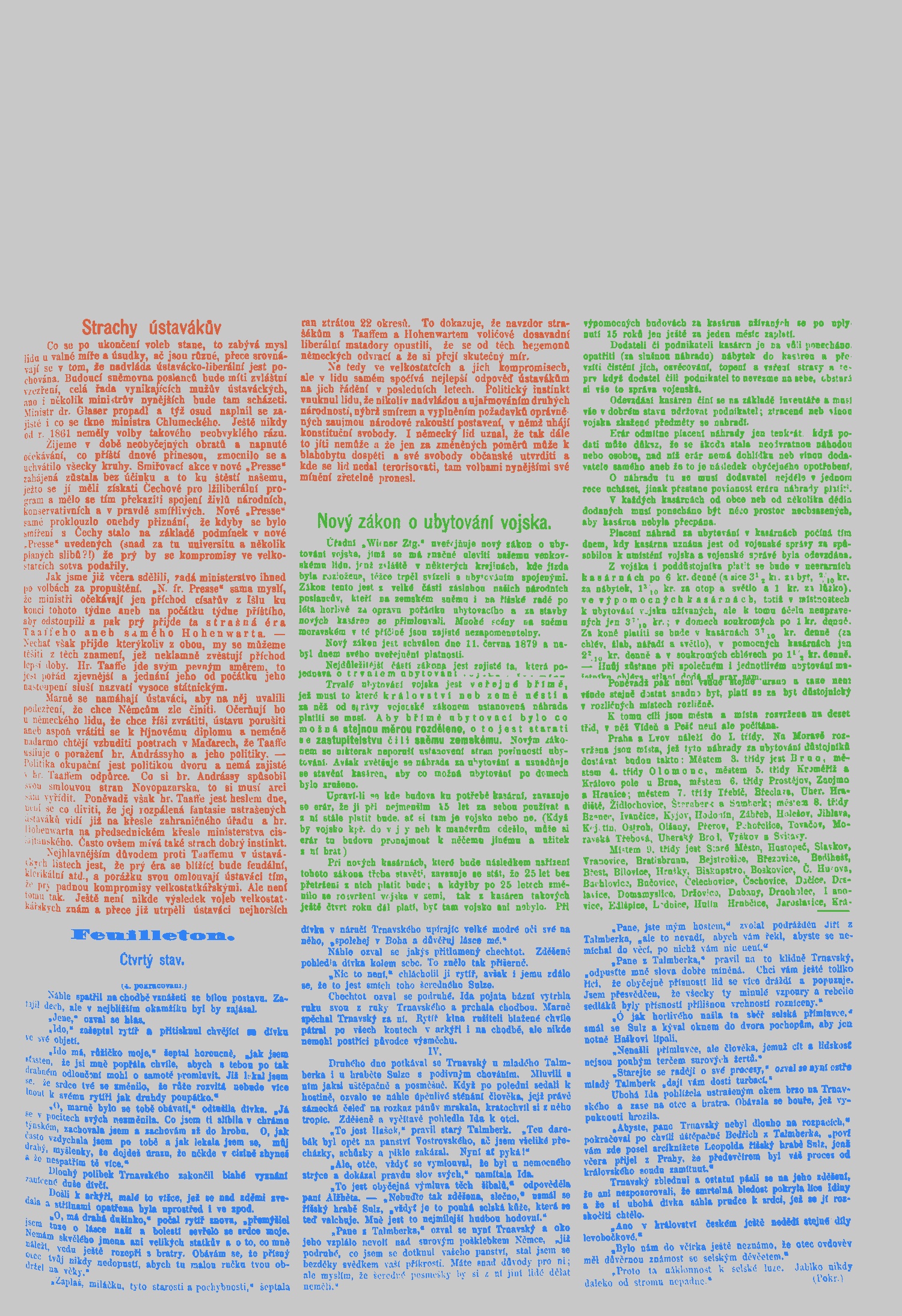}
            \caption{Logical segmentation}
            \label{fig:subfig2}
        \end{subfigure}
        \caption{Example of logical segmentation of a newspaper page as proposed in TextBite. Each segment is denoted in a different color. Only the colored pixels are considered in the evaluation. Page metadata, such as the title or edition are not included in the segmentation, as they are not a thematically coherent segments. However, our evaluation scheme does not penalize marking these parts of the page as additional segments.}
        \label{fig:segmentation-example}
    \end{figure}

    In this work, we propose evaluating logical page segmentation as a clustering problem. Rather than performing evaluation at the text level or relying on detection-based metrics, we treat segments as clusters of pixels within the document image. 
    Additionally, we argue that background pixels carry no semantic information relevant to the represented text and thus should be excluded from evaluation. 
    Foreground pixels, which correspond directly to individual letters, can be reliably identified using relatively simple segmentation methods, such as thresholding or binarization. 
    By considering only the pixels forming individual letters, our approach removes dependency on OCR accuracy, eliminates variations arising from different detection methods, and provides a unifying evaluation metric applicable to all types of logical page segmentation approaches.
    
    To support further research in this direction, we introduce a new dataset for logical page segmentation, based on historical Czech documents called TextBite. Our dataset comprises of 8,449 manually annotated pages with varying layouts and containing both printed and handwritten pages. This publicly available dataset can be used to develop and benchmark document segmentation systems.  An example of a page from the TextBite dataset together with its segmentation can be seen in Figure~\ref{fig:segmentation-example}.
    
    Furthermore, we created a set of baseline models, ranging from a traditional detection-based method to detection methods augmented with relation-prediction models, to establish baseline results and demonstrate the proposed evaluation framework.
    
    The main contributions of this paper can be summarized as follows:
    \begin{enumerate}
        \item We propose a novel evaluation approach to logical page segmentation based on pixel clustering, making it independent of OCR and insensitive to irrelevant geometric variations of detected text regions.
        \item We provide a hand-annotated dataset for logical page segmentation and other tasks, containing 8,449 pages.
        \item We provide a set of baseline solutions for this task along with their implementation and an evaluation framework.
    \end{enumerate}

\section{Related Work}

\subsection{Document Layout Datasets}
    Many datasets exist for layout analysis of both modern and historical documents; however, to our knowledge, no existing benchmarks employ an evaluation strategy similar to ours. Typically, benchmarks for these datasets frame evaluation either as a detection problem or as a relation-prediction problem and assess performance accordingly.

    Some datasets, such as PubLayNet \cite{publaynet} and DocLayNet \cite{pfitzmann2022doclaynet} frame layout analysis as a detection task, providing annotations for document elements like text, tables, and figures. Similarly, DocBank \cite{li2020docbank} offers large-scale OCR and detection annotations, making it a common benchmark for layout-based pretraining. Other datasets, such as CompHRDoc \cite{wang2024detect} and its superset HRDoc \cite{ma2023hrdoc}, go beyond detection by modeling hierarchical relationships, where elements are not only identified, but also classified and connected within a document’s structure. These datasets introduce parent-child relations between document components, making them particularly useful for tasks involving nested and multi-column structures, such as historical newspapers. Datasets like ROOR \cite{zhang2024modeling} focus on reading order prediction, evaluating how well models can reconstruct the logical sequence of content, which is crucial for multi-column or complex layouts.
    
    For historical documents, datasets such as Europeana Newspapers \cite{europeana}, and TexBiG \cite{texbig} introduce additional challenges related to the varying styles of printing, ornamental decorations, and manuscript-specific layouts. HBA 1.0 \cite{hba} provides pixel-level annotations for historical books, marking textual content and elements such as initials and illustrations.
    
    Despite the diversity of these benchmarks, none explicitly evaluate document segmentation as a pixel clustering task. Most existing approaches rely on bounding box detection, region classification, or relation modeling.

\subsection{Document Layout Analysis and Segmentation Methods}
    One way to achieve a logical segmentation is to first extract the textual content and then apply text segmentation methods. Older methods like TextTiling \cite{texttiling} or GraphSeg \cite{graphseg} use algorithmic techniques to detect boundaries in a continuous block of text. A more recent approach, proposed by Somasundaran et al. \cite{somasundaran2020two} utilizes two hierarchically connected transformer modules in a two learning objective setting. However, pure text-based approaches suffer in complex layouts by being dependent on reading order and OCR results.

    To overcome these shortcomings, segmentation at the page level is necessary, as it enables effective integration of visual and spatial features. 
    Early methods relied on convolutional neural networks for detecting layout elements~\cite{maskrcnn,liu2016ssd,ren2015faster,khanam2024yolov11}. 
    More recent approaches, such as DLAFormer \cite{wang2024dlaformer}, integrate object detection, information retrieval and reading order prediction within an end-to-end model, improving both the detection accuracy and relation prediction. 
    Alternatively, Wang et al.~\cite{wang2024detect} introduced a three-stage framework that constructs a hierarchical document structure, facilitating information retrieval from segmented pages.
    \sloppy
    Pre-trained transformer models are also widely employed in this domain~\cite{da2023vision,tu2023layoutmask,powalski2021going}. LayoutLM and its successors~\cite{xu2020layoutlm,xu2020layoutlmv2,huang2022layoutlmv3} integrate spatial embeddings with text tokens. \fussy 
    Document Image Transformer (DiT)~\cite{li2022dit} introduces a self-supervised training approach using large-scale unlabeled images. 
    Similarly, DocFormer~\cite{appalaraju2021docformer} is trained in an unsupervised manner, using text, vision, and spatial information within a multi-modal attention framework. 
    LiLT~\cite{wang2022lilt} takes a language-independent approach, allowing pre-training in one language with subsequent fine-tuning in another.

    Lastly, several pixel level segmentation methods have also been developed. 
    Sang et al.~\cite{seg1} proposed a pixel-wise spatial attention module designed to capture long distance pixel dependencies and they employed a conditional random field to model contextual information. 
    DocSegTr~\cite{biswas2022docsegtr}, on the other hand, treats the task as instance-level segmentation, proposing an end-to-end transformer model enhanced with a twin attention module.

\section{TextBite Dataset}
\begin{figure}[htp]
        \centering
        \begin{subfigure}{0.48\textwidth}
            \centering
            \includegraphics[width=\linewidth]{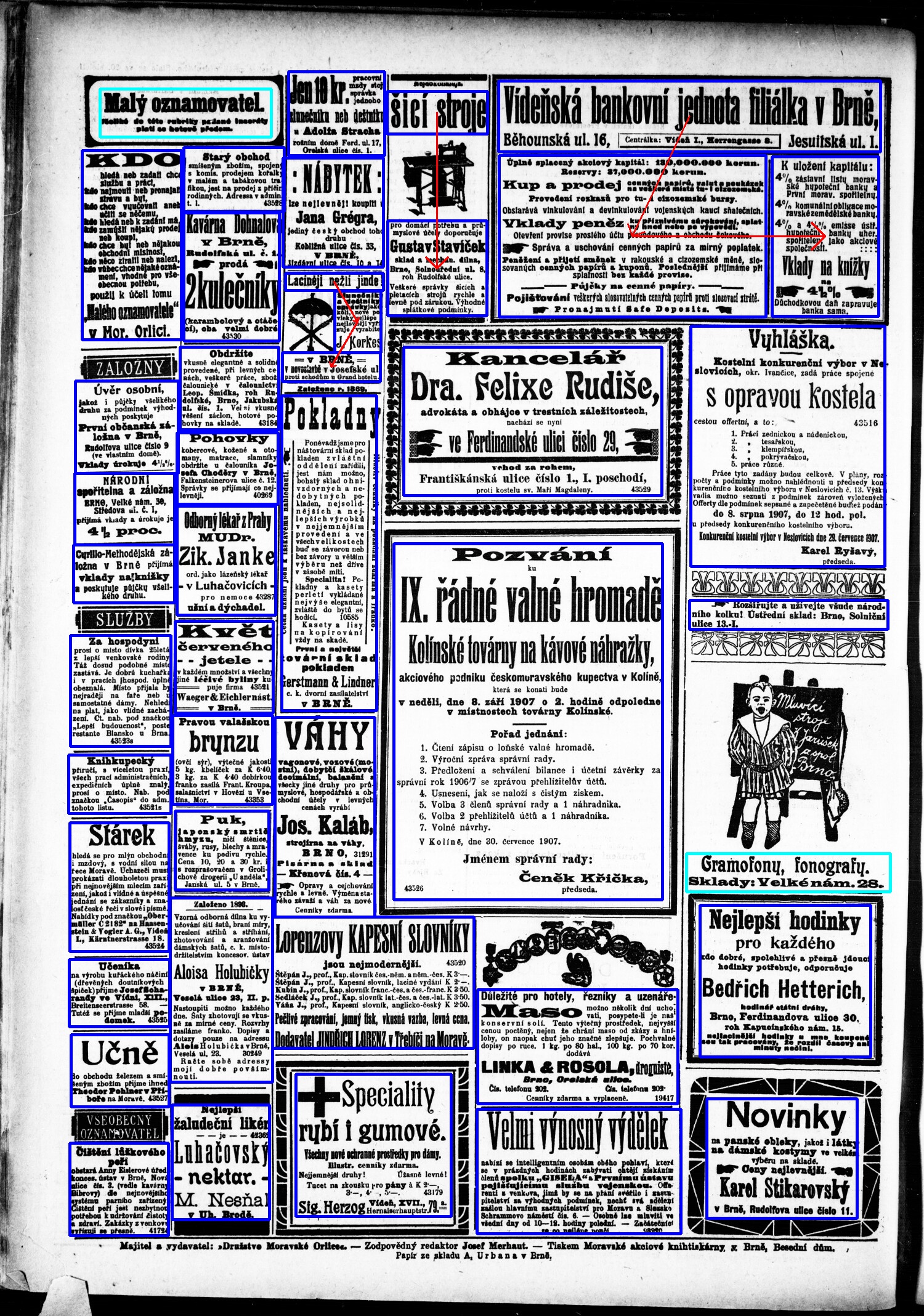}
            \caption{Newspaper advertisements}
            \label{fig:sf:example1}
        \end{subfigure}
        \begin{subfigure}{0.48\textwidth}
            \centering
            \begin{minipage}{\textwidth}
                \centering
                \includegraphics[width=\linewidth]{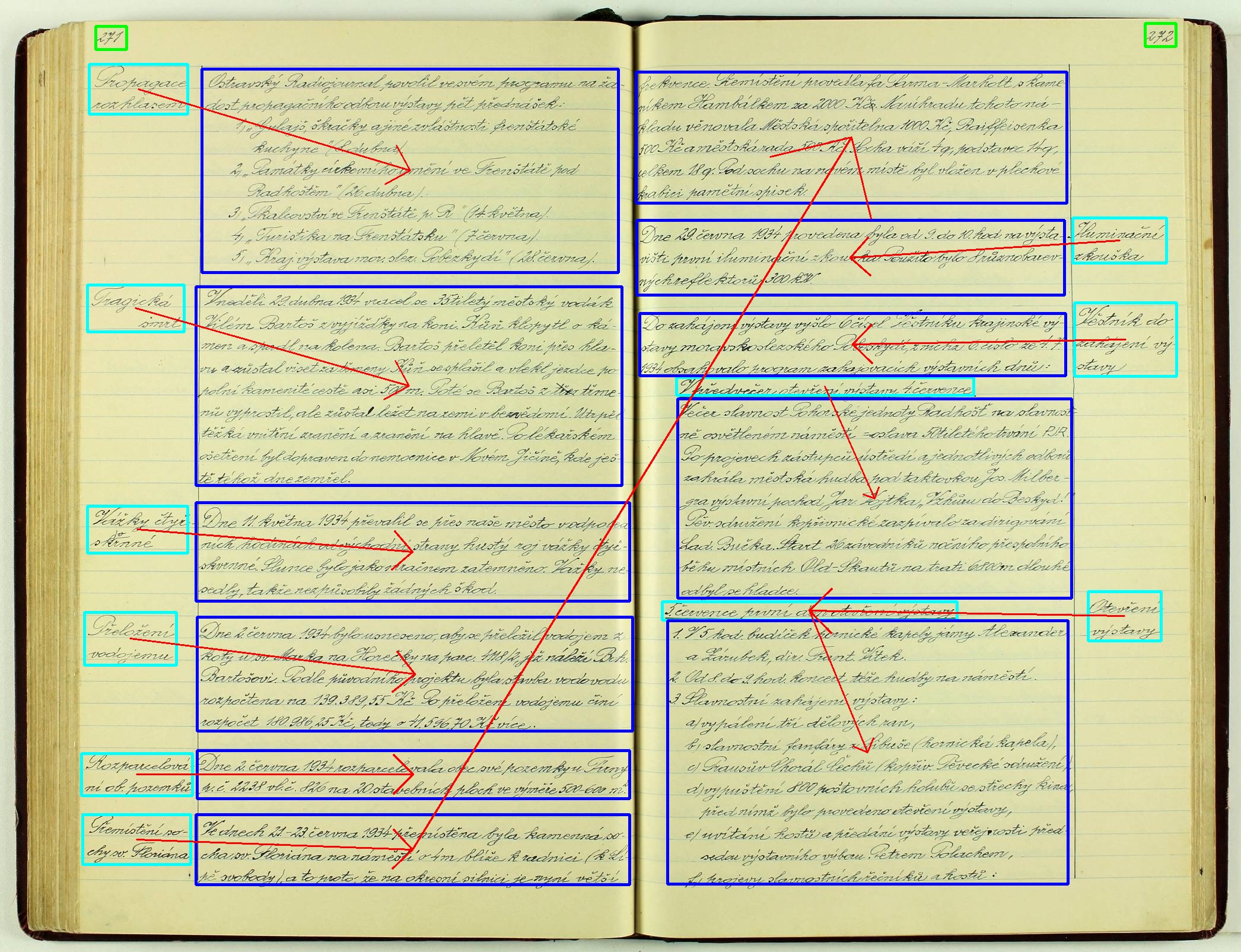}
                \includegraphics[width=\linewidth]{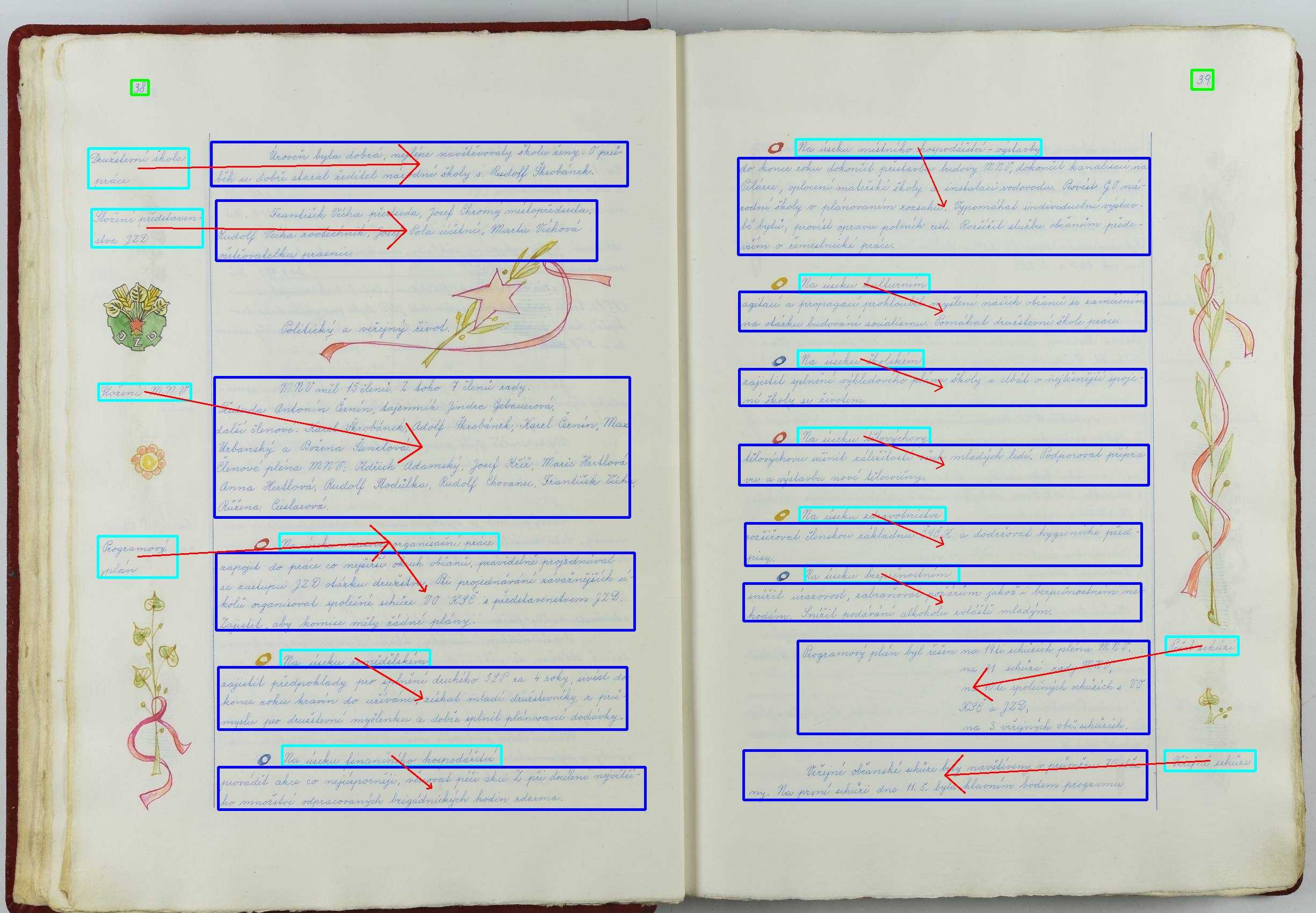}
            \end{minipage}
            \caption{Two handwritten pages}
            \label{fig:sf:example2}
        \end{subfigure}
        \begin{subfigure}{0.48\textwidth}
            \centering
            \includegraphics[width=\linewidth]{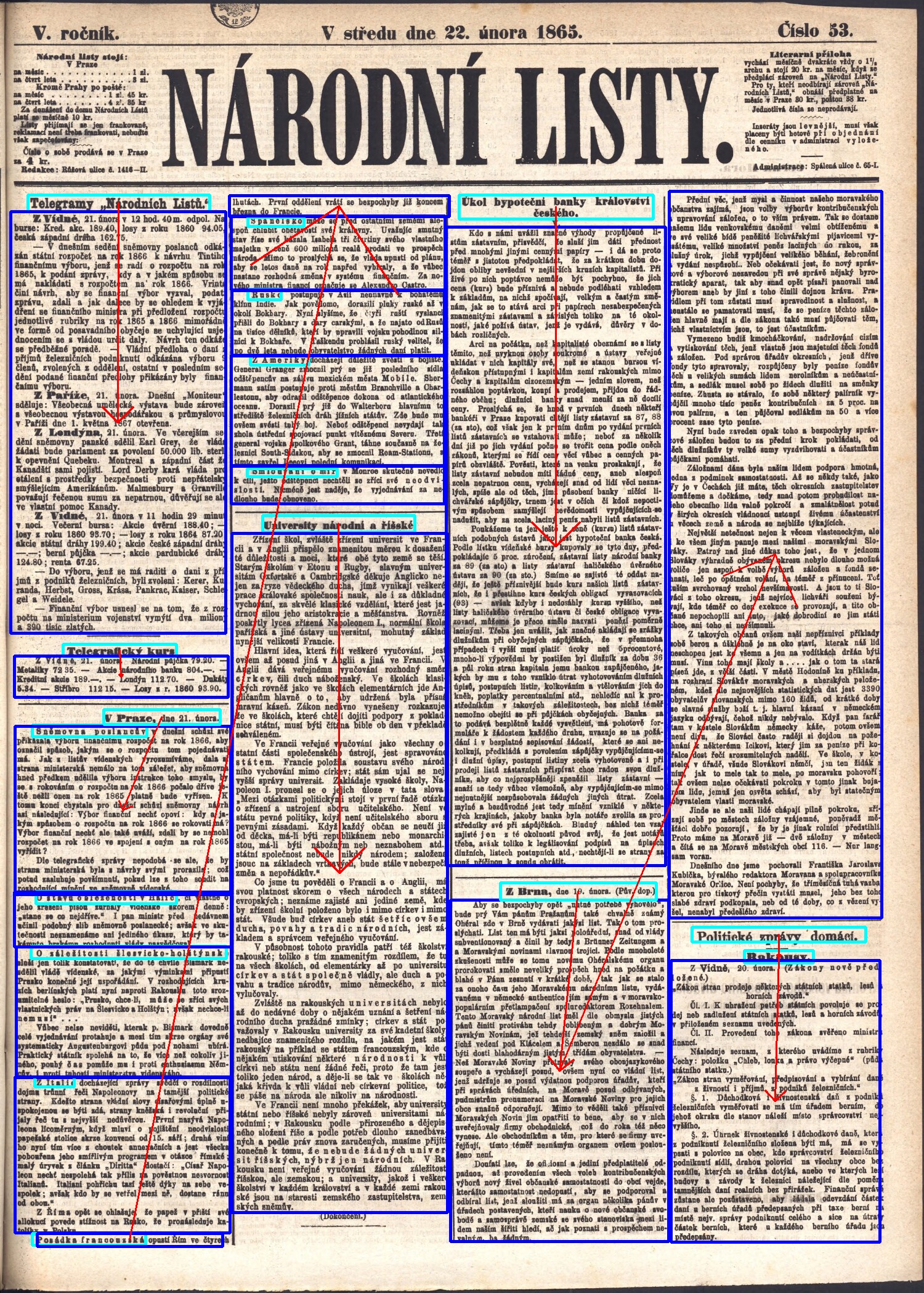}
            \caption{Newspaper title page}
            \label{fig:sf:example3}
        \end{subfigure}
        \begin{subfigure}{0.48\textwidth}
            \centering
            \includegraphics[width=\linewidth]{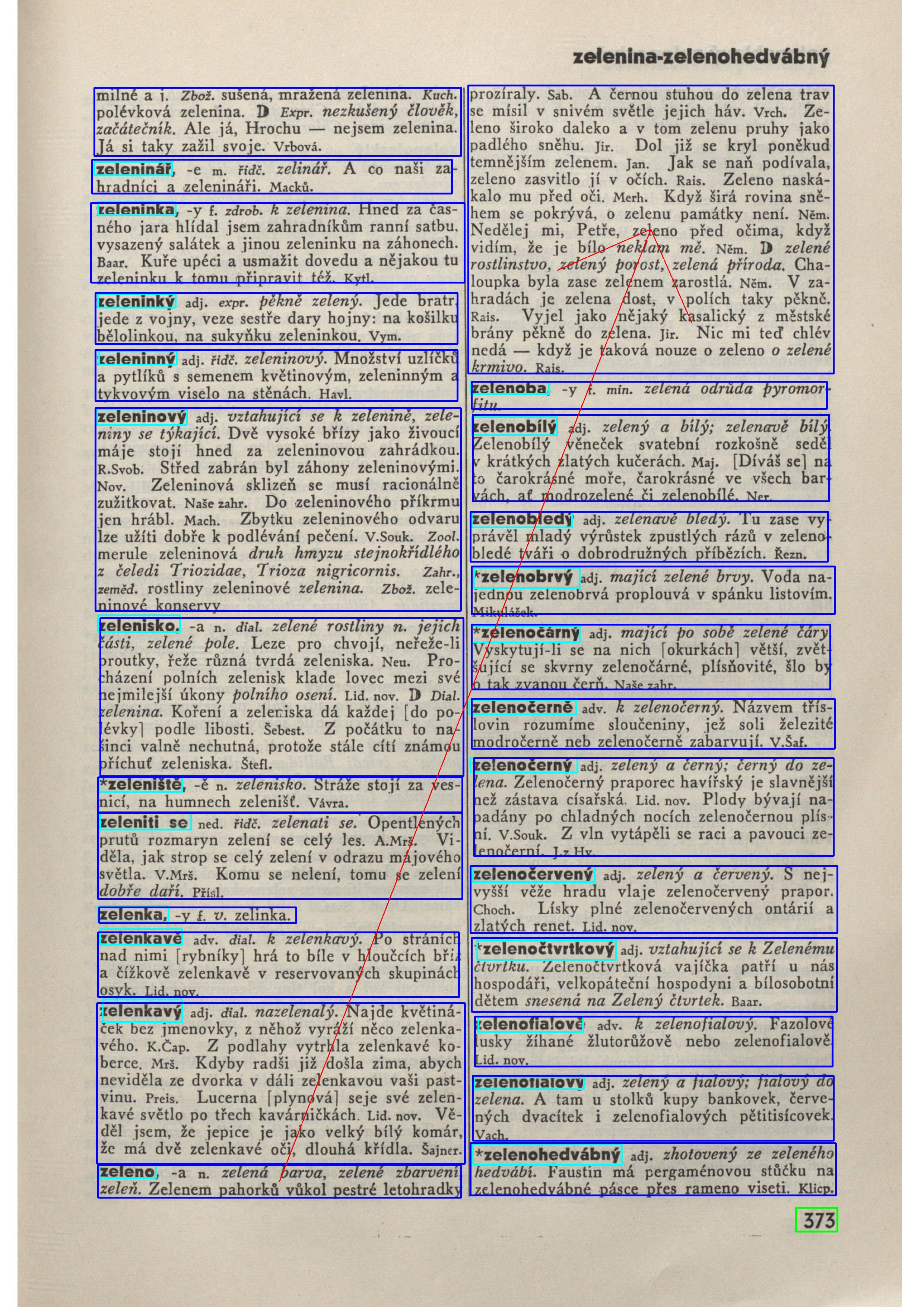}
            \caption{Dictionary page}
            \label{fig:sf:exampl4}
        \end{subfigure}
    
        \caption{Examples of annotated pages in in TextBite dataset with various layouts.}
        \label{fig:annotation-example}
    \end{figure}

    The TextBite dataset consists of scanned historical Czech documents from various sources with diverse layouts. 
    It includes simpler layouts, such as book pages and dictionaries, as well as more complex multi-column formats from newspapers, periodicals, and other printed materials. 
    Additionally, part of the dataset contains handwritten documents, primarily records from schools and public organizations, introducing extra segmentation challenges due to their more loosely structured layouts. Examples of pages from the dataset are visualized in Figure \ref{fig:annotation-example}.
    In total, the dataset contains 8,449 annotated pages, from which 7,346 pages are printed and 1,103 are handwritten. The pages contain a total of 78,863 segments. To ensure consistency in the evaluation, we split the dataset into a development and test subset. The test subset contains 964 pages, of which 185 are handwritten. The annotations are provided in an extended \texttt{COCO} format \cite{coco}.
    
    In the annotations, each segment is represented by a set of axis aligned bounding boxes, which are connected by directed relationships, representing reading order. To include these relationships in the \texttt{COCO} format, a new top-level key \texttt{relations} is added. Each relation entry specifies a source and a target bounding box. Subsequently, this style of annotations implicitly creates an intra-segment reading order.
    
    \sloppy
    In addition to the layout annotations, we provide a textual representation of the pages produced by Optical Character Recognition (OCR) tool \texttt{PERO-OCR}~\cite{pero1,pero2,pero3}. These come in the form of XML files in the \texttt{PAGE-XML}~\cite{pagexml} format, which includes an enclosing polygon for each individual textline along with the transcriptions and their confidences. Lastly, we provide the OCR results in the \texttt{ALTO}~\cite{alto} format, which includes polygons for individual words in the page image.\fussy

    Apart from the proposed logical page segmentation, the dataset can be used in multiple research tasks, such as reading order prediction, title detection and association or document layout analysis in general.

\subsection{Data Acquisition}
    Several Czech libraries are currently digitizing historical documents to preserve national heritage, making most of these documents publicly accessible for personal and research use. Librarians actively support this effort by organizing, sorting, and filtering the documents, creating collaboration opportunities between libraries and public institutions, such as universities.

    As a result, these document collections became the primary data source for the TextBite dataset. 
    The printed document pages were sampled through multiple iterations. Initially, around 4,000 pages spanning various document types and historical periods were manually selected from a single library. 
    Then, approximately 1,000 newspaper title pages were selected and further filtered by annotators. 
    The remaining printed pages were randomly sampled from several libraries without specific criteria regarding document types but were carefully filtered to ensure suitability for segmentation tasks.

    We specifically selected only pages written in the Czech language, although occasional traces of other languages appear. 
    German is the most prevalent secondary language, while languages such as French occur only occasionally—for example, as dictionary keys in translation dictionaries.
    Filtering also excluded pages that were damaged, rotated, overly blurry, or otherwise unsuitable for segmentation.
    
\subsection{Data Annotation}
\begin{figure}[ht]
        \centering
            \includegraphics[scale=0.65]{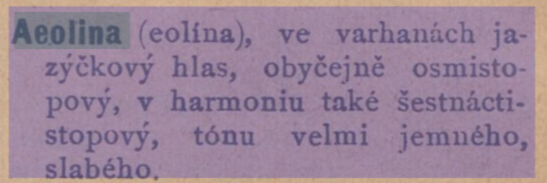}
        \caption{Cropped section of an annotated image showcasing an enclosed title region (green) in a text region (blue).}
        \label{fig:enclosed-title}
    \end{figure}

    \noindent\textbf{Annotation Structure.} Each page contains two types of annotations: bounding boxes and relations. Bounding boxes enclose a single-column text region and are categorized as text region, title, or page number. Page number annotations were created at a later stage as an additional sub-task, but in the eyes of logical segmentation, they are excluded from the evaluation. Relations establish coherent segments by linking bounding boxes, forming an implicit reading order. Titles are connected to their corresponding text blocks unless fully enclosed in other region (see Figure~\ref{fig:enclosed-title}). This happens in dictionaries and newspapers, where the title is included in the text and usually begins with it (named paragraphs). In such case, the title annotation has no effect on the evaluation. Note that the page numbers have no relations. 

    We did not annotate any page metadata, as it does not contribute to the core informational content of the page. Elements such as headers, footers, and other details of the publication were excluded to keep the focus on meaningful text. For example, in newspapers, details such as the publication date, newspaper name, or edition were not annotated.

    \noindent\textbf{Annotation Process.} The annotation process was conducted using the open-source data labeling tool \texttt{Label Studio}~\cite{labelstudio}. After setting up the labeling interface, we prepared a set of instructions to ensure consistency in the annotations. Additional, more detailed guidelines and feedback were given to annotators during initial training. More than ten people participated in the annotation process, including university students, librarians, and researchers. Most of the annotation work was carried out by students and librarians. Researchers served mostly as supervisors. After the annotations were completed, researchers manually checked and corrected the test portion of the dataset, whereas students reviewed the development portion.

\subsection{Dataset Characteristics}
\begin{figure}[ht]
        \centering
        \begin{subfigure}{0.49\textwidth}
            \centering
            \includegraphics[width=\linewidth]{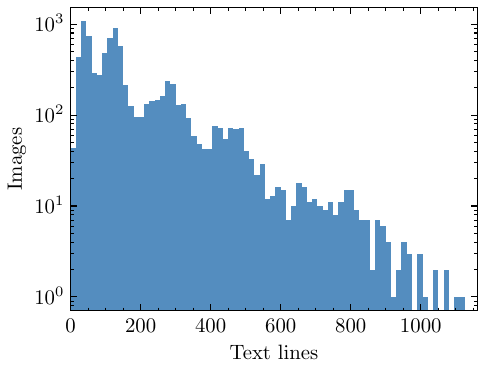}
            \caption{Textline count on pages}
            \label{fig:subfig1:textline-count}
        \end{subfigure}
        \hfill
        \begin{subfigure}{0.49\textwidth}
            \centering
            \includegraphics[width=\linewidth]{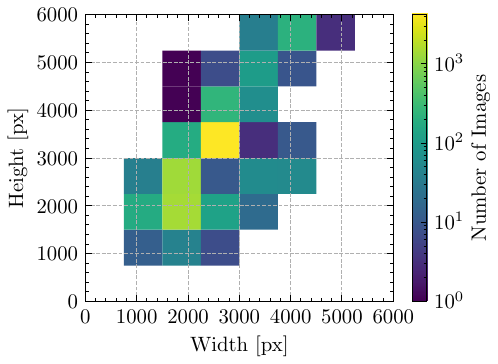}
            \caption{Image dimensions}
            \label{fig:subfig2:image-dimensions}
        \end{subfigure}
        \begin{subfigure}{0.49\textwidth}
            \centering
            \includegraphics[width=\linewidth]{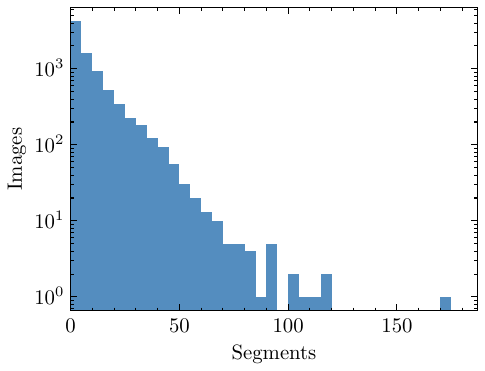}
            \caption{Segment count on pages}
            \label{fig:subfig3:segment-count}
        \end{subfigure}
        \hfill
        \begin{subfigure}{0.49\textwidth}
            \centering
            \includegraphics[width=\linewidth]{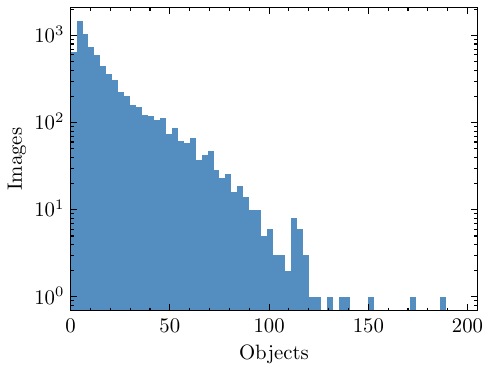}
            \caption{Object count on pages}
            \label{fig:subfig4:object-count}
        \end{subfigure}
        \caption{Data characteristics of the TextBite dataset.}
        \label{fig:dataset-characteristics}
    \end{figure}

    The TextBite dataset presents a diverse range of historical Czech documents with varying complexity in their layouts. Table \ref{table:object-count-category} provides an overview of the number of annotated regions in different categories for both printed and handwritten documents. Notably, printed documents usually contain a higher number of text regions and titles, while handwritten documents contain relatively fewer, but often more complicated, text segments. Figure \ref{fig:dataset-characteristics} presents an overview of the dataset's characteristics, including the number of textlines, image dimensions, segment count per image, and region count per image.

\begin{table}[ht]
	\centering
	\caption{Number of annotated regions of their respective class in both printed and handwritten pages.}
	\label{table:object-count-category}
	\begin{tabular}{@{\hspace{0.1cm}}l@{\hspace{0.3cm}}c@{\hspace{0.3cm}}c@{\hspace{0.1cm}}}
		\toprule	
		Category & Printed & Handwritten \\
		\midrule
		Title & 46,140 & 2,648 \\
		Text & 76,469 & 3,989 \\
		Page number & 5,248 & 965 \\
		\bottomrule
	\end{tabular}
\end{table}

\subsection{Evaluation}
    The TextBite dataset is annotated with linked bounding boxes that define the logical segments. While the annotations were created with reasonable precision, the exact placement of box boundaries is irrelevant to the evaluation. Instead, the evaluation process focuses on textual content within the annotated regions rather than on their precise spatial boundaries.

    This is achieved by producing a per-pixel mask, which describes where the individual logical segments are. Pixels not belonging to any segment are then excluded from the evaluation, giving system designers freedom in dealing with the ambiguous regions in-between text regions and making it easier to train systems on weakly supervised data where precise boundaries are not available.
    \begin{figure}[ht]
        \centering
        \begin{subfigure}{0.31\textwidth}
            \centering
            \includegraphics[width=\linewidth]{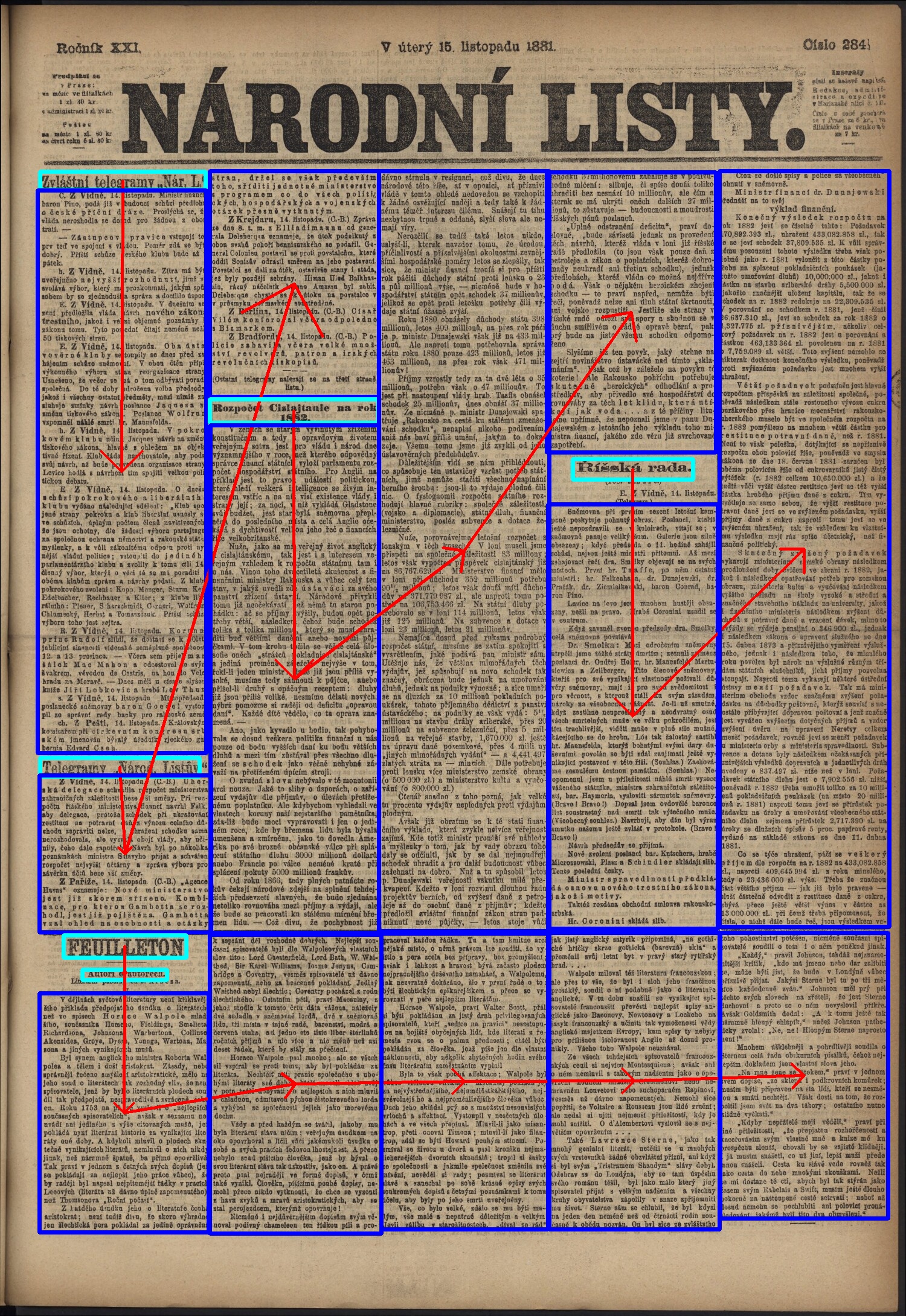}
            \caption{%
                Human annotation
            }\label{fig:subfig1:page-from-label-studio}
        \end{subfigure}
        \hfill
        \begin{subfigure}{0.31\textwidth}
            \centering
            \includegraphics[width=\linewidth]{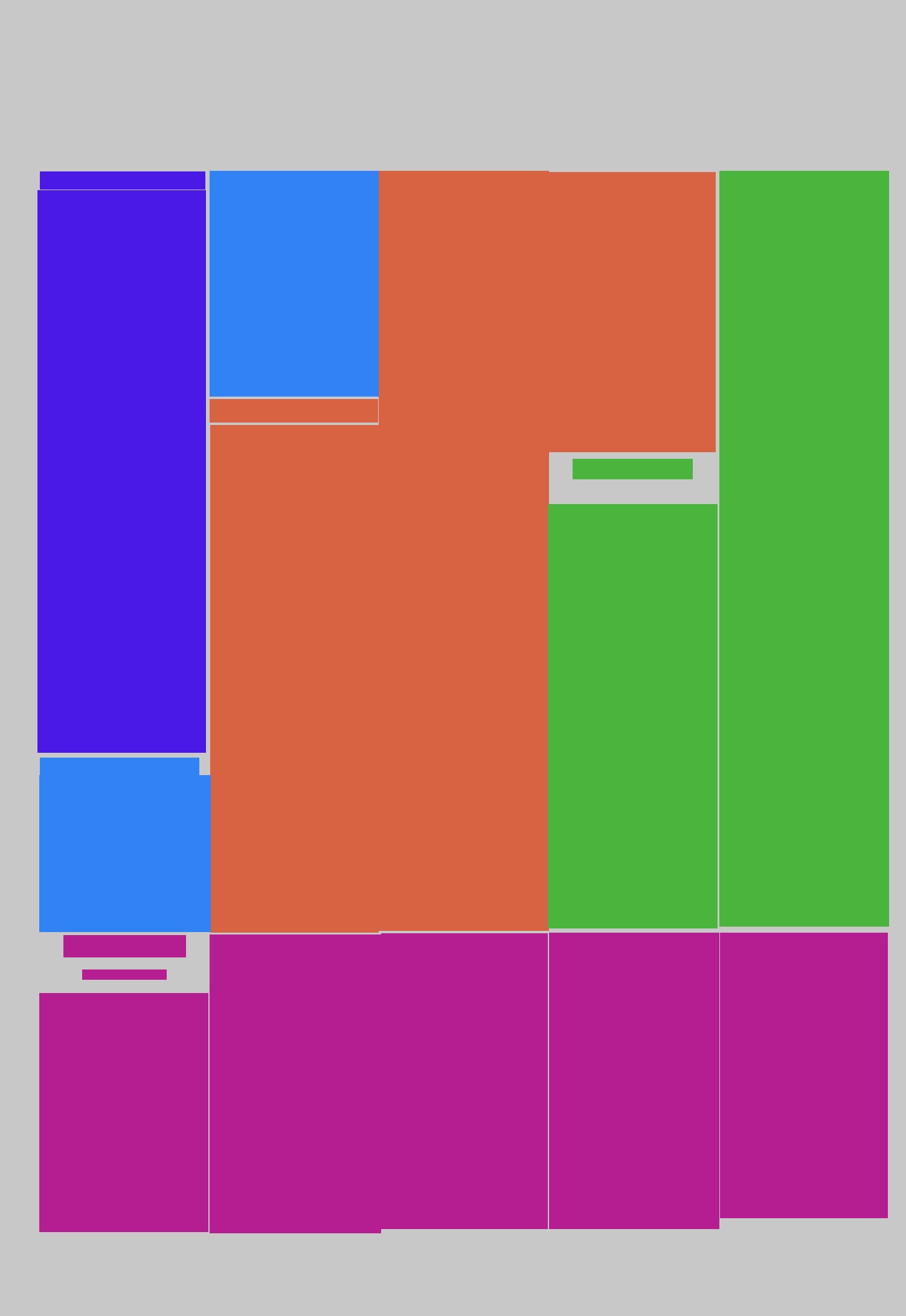}
            \caption{%
                Annotation mask
            }\label{fig:subfig2:gold-annotation}
        \end{subfigure}
        \hfill
        \begin{subfigure}{0.31\textwidth}
            \centering
            \includegraphics[width=\linewidth]{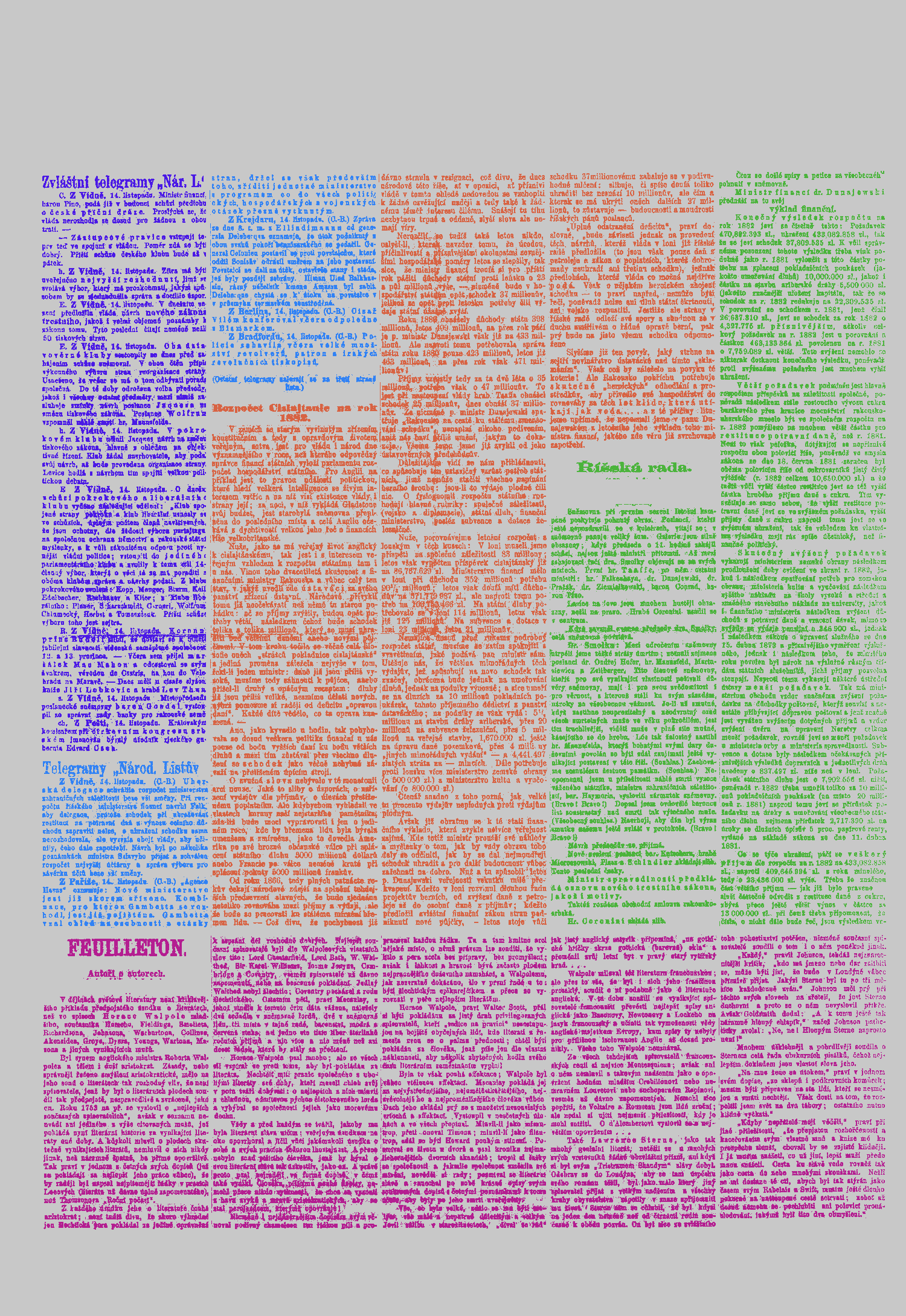}
            \caption{
                Pixel segmentation
            }\label{fig:subfig3:pixel-segmentation}
        \end{subfigure}
        \caption{%
            Construction of the ground truth pixel segmentation of a page.
            First, we turn the human annotation in the form of connected regions~\ref{fig:subfig1:page-from-label-studio} into their masked representation~\ref{fig:subfig2:gold-annotation}. This is then intersected with textlines obtained from OCR and thresholded, which results in the final pixel segmentation~\ref{fig:subfig3:pixel-segmentation}.
        }\label{fig:evaluation-construction}
    \end{figure}

    We produce the pixel annotations by narrowing the output from annotators in two ways:
    First, we intersect the manually annotated regions with textlines as detected by the PERO-OCR.
    This way, we constrain the pixel annotations so that larger areas of empty space do not contribute to the definitions of logical segments, which happens e.g.\ when the text is not justified.
    Second, we use thresholding to further limit the extent of the pixel annotation.
    Specifically, we use adaptive thresholding, with the threshold value estimated dynamically from a patch of 301x301 pixels.
    Using such fine segmentation, we (a) further prune areas without text and (b) allow direct evaluation of systems operating at the pixel level.

    Finally, we evaluate the segmentation output using the Rand index \cite{rand}. Since the evaluation focuses only on text regions, we compute the Rand index over pixels within the ground truth pixel segmentation. For each pair of pixels, we mark if they are in the same segment in the ground truth and in the system output. The Rand index is then the proportion of pixel pairs where these assignments agree, with its value ranging from 0 (no agreement) to 1 (perfect agreement). If a pixel is assigned to a segment in the ground truth but is missing in the system output, it is treated as a misassignment, reducing the score. In practice, when 10~\% of the considered pixels are misassigned, the Rand index is typically around 0.83, with the exact value depending on the specific misassignments.

    Since pixels marked as background in the ground truth are not taken into consideration, segmentation systems are free to mark page numbers, page-level titles etc.\ as additional segments without hurting their Rand index.
    On the other hand, if the system output fails to label a part of a ground truth segment, the corresponding pixels are treated as assigned to an extra cluster, decreasing the Rand index.

    While the exact pixel annotation arises from this automatic process and is thus not gold-label in a strict sense, the possible imperfections do not jeopardize the evalation scheme:
    Since the pixel annotations are derived from manually verified human annotations that accurately localize text, they do not introduce incorrect clusters.
    In the case when the textline detection or thresholding act too aggressively and omit a portion of the text, the result is just a decreased weight of the specific segment, which still allows for a fair comparison between different systems.

\section{Baseline Experiments}
    To establish a reference point for future research and facilitate further development, we have implemented a set of baseline methods. These methods serve as benchmarks, allowing for an objective evaluation of performance and providing insights into the challenges posed by the task. By assessing the effectiveness of these baselines, we aim to create a foundation for future improvements and encourage the development of more advanced approaches.

    The development portion of the dataset was split into a training and validation set. The validation set contains 1,000 images, of which 200 are handwritten. The remaining 6,400 images were used to train the models. To estimate the uncertainty of evaluation results, we computed 95\% confidence intervals using bootstrapping.
    
\subsection{YOLO Detection Model}
    The first baseline method, which also serves as the basis for the other methods, detects regions of text using the YOLOv11~\cite{khanam2024yolov11} object detector finetuned to the training set. This model identifies bounding boxes on a page, which represent distinct segments. To ensure consistency, any detected bounding box that is fully enclosed within a larger bounding box of is filtered out.
    
    To train the detector, we used the publicly available Ultralytics \cite{Jocher_Ultralytics_YOLO_2023} implementation. We experimented with three model variants and different image resolutions, with the results summarized in Table~\ref{table:yolo-training}. Ultimately, we selected the small variant with a resolution of 1,200 pixels. All models were trained with a batch size of 16, a learning rate of $2e^{-4}$, and the Adam optimizer. Training was terminated after 20 epochs of no improvement on the validation set. The best-performing model, evaluated on the test data, achieved a Rand index of 83.9~\%  with a 95\% confidence interval of $\left[82.6, 85.1\right]$.
    
    This pure detection approach has a significant limitation. The model treats all predicted regions as independent segments, whereas, in reality, a segment may span multiple columns, include an associated title positioned above the main text, or it may consist of multiple bounding boxes. To address this, we introduce a second stage into the pipeline designed to merge detected bounding boxes into coherent segments.
    \begin{table}[ht]
    	\centering
    	\caption{Training results of YOLOv11 object detection models in three sizes with five different resolutions. The best value in each category is highlighted.}
    	\label{table:yolo-training}
    	\begin{tabular}{@{\hspace{0.1cm}}l@{\hspace{0.3cm}}|@{\hspace{0.3cm}}ccccc@{\hspace{0.3cm}}|@{\hspace{0.3cm}}ccccc@{\hspace{0.1cm}}}
    		\toprule
            Metric & \multicolumn{5}{c}{mAP50} & \multicolumn{5}{c}{mAP50-95} \\
    		Resolution & 640  & 800 & 1000 & 1200 & 1400 & 640  & 800 & 1000 & 1200 & 1400 \\
    		\midrule
    		YOLOv11n & 86.4 & 89.1 & 90.7 & 91.3 & 91.5 & 65.4 & 68.2 & 69.9 & 70.9 & 70.8 \\
    		YOLOv11s & 88.0 & 90.0 & 91.2 & \textbf{92.0} & - & 67.9 & 69.9 & 71.4 & \textbf{72.2} & - \\
    		YOLOv11m & 90.0 & 91.4 & - & - & - & 70.3 & 71.4 & - & - & - \\
    		\bottomrule
    	\end{tabular}
    \end{table}

\subsection{Graph Neural Network}
    The first approach to detection merging is based on a graph neural network trained as a binary edge classifier. Each edge in the graph represents a potential connection between two detected regions, with the label indicating whether they should be merged.
    
\begin{figure}[ht]
        \centering
            \includegraphics[width=\textwidth]{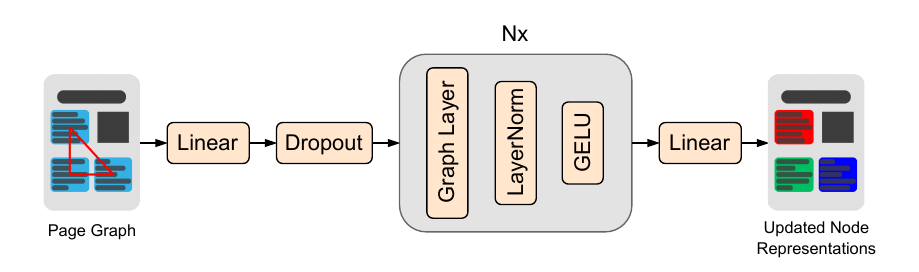}
        \caption{Graph neural network structure}
        \label{fig:gnn}
    \end{figure}

    Graphs are constructed as complete undirected graphs, where each node represents a detected region. During annotation, relationships were only defined between consecutive regions rather than all regions within a segment. To ensure that all detections belonging to a single segment are merged, we assigned positive labels to all edges within the same segment. Edge classification is performed by computing the cosine similarity between the updated node representations and comparing it to a classification threshold. The overall model architecture is illustrated in Figure \ref{fig:gnn}.
    
    The input features include 29 geometric node features and 12 edge features, which combine both geometric and text features. Nodes contain only geometric features representing the position of the region both in absolute and relative values. Geometric edge features capture spatial relationships, including absolute and relative distances between the centers of the region and the closest edges. Text-based features are created by computing a cosine similarity between text embeddings of the two connected regions. The text embeddings are obtained using the CZERT~\cite{sido2021czert} masked language model.
    
    Since both absolute and relative features are used, their scales vary significantly. To address this, we apply Z-score normalization, using mean and standard deviation values computed from the training set to normalize the data.

    We experimented with various model configurations. The best-performing model was trained with the Residual Gated Graph Convolution operator~\cite{bresson2017residual}, a batch size of 16, dimensionality of 512, 3 graph layers and AdamW optimizer with $1e^{-3}$ learning rate. On the test data, the model achieved a Rand index of 92.5 \% with a 95\% confidence interval $\left[91.6, 93.3\right]$.

\subsection{Transformer}
    The second region merging method is based on a transformer architecture with the task set up as an exclusive relation prediction. The input queries $Q = \{q_1, q_2, ..., q_n\}$ are the detected bounding boxes represented by 2D sinusoidal positional embedding vectors. Then, relation scores are calculated using the model output feature vectors $O = \{o_1, o_2, ... , o_N\}$ as:
    \[
        s_{i,j} = FC^D_a\left(o_i\right) \cdot FC^D_b\left(o_j\right); i \in \{1 .. N\}, j \in \{1 .. N\},
    \]
    where $FC^D$ is a fully-connected layer with dimensionality $D$. The highest scoring query $q_j$ is then chosen as the relation for query $q_i$. The relations are used to form chains of regions, which are then grouped together to create the final segments.

    In its basic form, the model consists of a transformer encoder. We also experimented with the usage of images as additional information for the model. In this case, a transformer decoder with parallel decoding is used. The image is first resized to the resolution of $1024 \times 1024$ and then processed by an image backbone and the resulting feature vectors are used in the cross-attention layer of the decoder. The results of these experiments can be seen in Table \ref{table:transformer-ablation}. We chose ResNet-18 \cite{he2016deep} and ResNet-50 as the backbones and we also experimented with freezing their weights during training. Adding the visual information increases the performance, but no significant difference can be observed between the ResNet variants and frozen weights. Due to these results, we chose the model with the frozen ResNet-18 image backbone as the final model.
    
    The models were trained with 6 layers, hidden dimension of 128, 2 attention heads, batch size of 4, learning rate of $2e^{-4}$ and the inputs were padded to 80 tokens. The cross-entropy loss function was used to train the model, acting on the relation scores. On the test data, the model achieved a Rand index of 86.9 with a  95\% confidence interval $\left[85.9, 88.0\right]$.
    \begin{table}[ht]
    	\centering
    	\caption{Transformer results on the validation data. Including images in the model improves its performance, but larger image backbone or freezing its weights has no effect on the model performance.}
    	\label{table:transformer-ablation}
    	\begin{tabular}{@{\hspace{0.1cm}}l@{\hspace{0.3cm}}c@{\hspace{0.3cm}}c@{\hspace{0.3cm}}c@{\hspace{0.1cm}}}
    		\toprule
    	    Image Backbone & Frozen & Rand Index & CI@0.95 \\
    		\midrule
    		None & - & 86.5 &$\left[85.4, 87.5\right]$ \\
    		ResNet-18 & No & 88.3 & $\left[87.3, 89.2\right]$ \\
    		ResNet-18 & Yes & 88.5 & $\left[87.5, 89.5\right]$ \\
            ResNet-50 & No & 87.5 & $\left[86.5, 88.5\right]$ \\
    		ResNet-50 & Yes & 88.9 & $\left[88.0, 89.8\right]$ \\
    		\bottomrule
    	\end{tabular}
    \end{table}

\subsection{Results Discussion}
    The results in Table \ref{table:final-results} show that the pure YOLOv11 detection model achieves a reasonable baseline performance, but its inability to detect more complicated segments limits its potential. Incorporating a Graph Neural Network significantly improves the results, achieving the highest performance with a 95\% confidence interval of $\left[91.583, 93.291\right]$. The transformer-based approach, while outperforming the standalone YOLOv11 model, does not reach the performance of the GNN. Additionally, using image features in the transformer improved performance, indicating that visual context can be beneficial.
    \begin{table}[ht]
    	\centering
    	\caption{Final results of all the proposed methods on the test portion of the TextBite dataset.}
    	\label{table:final-results}
    	\begin{tabular}{@{\hspace{0.1cm}}lcc@{\hspace{0.1cm}}}
    		\toprule
    	    Method & Rand Index & CI@0.95 \\
    		\midrule
    		YOLOv11 & 83.9 & $\left[82.6, 85.1\right]$ \\
    		GNN & 92.5 & $\left[91.6, 93.3\right]$ \\
    		Transformer & 86.9 & $\left[85.9, 88.0\right]$ \\
    		\bottomrule
    	\end{tabular}
    \end{table}

\section{Conclusions}
    In this paper, we introduced the TextBite dataset for logical page segmentation and document layout analysis. It consists of 8,449 manually annotated pages from historical Czech documents, encompassing a diverse range of layouts from periodicals, books, and handwritten manuscripts. The dataset includes annotations in an extended \texttt{COCO} format along with transcriptions generated by \texttt{PERO-OCR}. It is publicly available online\footnote{The dataset is available at: \url{https://github.com/DCGM/textbite-dataset}.}, providing a valuable resource for research in logical document segmentation and layout analysis.

    To address the limitations of existing segmentation evaluation methods, we proposed a novel approach that formulates logical page segmentation as a pixel clustering problem. By representing segments as clusters of foreground pixels corresponding to individual letters, our approach eliminates the dependency on OCR accuracy and mitigates the impact of geometric variations in detected text regions. This unifying evaluation framework enables fair comparisons across different segmentation methods.
    
    To establish baselines, we implemented and evaluated three methods: a traditional object detector, region merging with a graph neural network, and region merging with a transformer model. The best-performing method—YOLOv11 combined with a graph neural network—achieved a Rand index of 92.5\%. These results highlight the potential of our evaluation framework and dataset to advance research in document segmentation and layout analysis.

\begin{credits}
\subsubsection{\ackname} This work has been supported by the Ministry of Culture Czech Republic in NAKI III project semANT - Semantic Document Exploration (DH23P03OVV060).

\subsubsection{\discintname}
The authors have no competing interests to declare that are relevant to the content of this article.
\end{credits}
%
% ---- Bibliography ----
%
% BibTeX users should specify bibliography style 'splncs04'.
% References will then be sorted and formatted in the correct style.
%
% \bibliographystyle{splncs04}
% \bibliography{mybibliography}
%
\bibliographystyle{splncs04}
\bibliography{refs}

\end{document}